\documentclass[conference,10pt,letterpaper,twoside]{style/ieeeconf}

\IEEEoverridecommandlockouts    
\makeatletter 

\let\proof\@undefined
\let\endproof\@undefined

\let\NAT@parse\undefined
\makeatother

\usepackage[numbers, sort, compress]{natbib}
\usepackage{graphicx}
\usepackage{amsmath,amssymb}
\usepackage{caption}
\usepackage{tabularx}
\usepackage[font=footnotesize,labelsep=period]{caption}
\usepackage{subcaption}
\usepackage{soul}
\usepackage{hyperref} 
\usepackage{url} 
\usepackage{multirow}
\usepackage{footmisc}

\usepackage{amsthm}
\usepackage{xcolor}

\usepackage{style/perl_acronyms}
\usepackage{style/perl_math}
\usepackage{style/perl_SIunits}
\usepackage{style/perl_misc}

\usepackage{stfloats}

\begin{document}

\title{Terra: Hierarchical Terrain-Aware 3D Scene Graph for Task-Agnostic Outdoor Mapping}

\author{Chad~R.~Samuelson, Abigail~Austin, Seth~Knoop, Blake~Romrell, \\ Gabriel~R.~Slade, Timothy~W.~McLain,~and Joshua~G.~Mangelson%
  \thanks{This work was partially funded under Office of Naval Research award numbers N00014-24-1-2301 and N00014-24-1-2503, as well as by the Center for Autonomous Air Mobility and Sensing (CAAMS), a National Science Foundation Industry-University Cooperative Research Center (IUCRC) under NSF award number 2139551, along with significant contributions from CAAMS industry members.}
  \noindent
  \thanks{C.~Samuelson, A.~Austin, S.~Knoop, B.~Romrell, G~Slade, T.~McLain, and J.~Mangelson are at Brigham Young University. They can be reached at \texttt{\{chadrs2, abiausti, sk8723, romrellb, grs45, mclain, mangelson\}@byu.edu}.}  %
}

\maketitle

\acresetall

\IEEEpeerreviewmaketitle

\begin{abstract}

Outdoor intelligent autonomous robotic operation relies on a sufficiently expressive map of the environment. Classical geometric mapping methods retain essential structural environment information, but lack a semantic understanding and organization to allow high-level robotic reasoning. 3D scene graphs (3DSGs) address this limitation by integrating geometric, topological, and semantic relationships into a multi-level graph-based map. Outdoor autonomous operations commonly rely on terrain information either due to task-dependence or the traversability of the robotic platform. We propose a novel approach that combines indoor 3DSG techniques with standard outdoor geometric mapping and terrain-aware reasoning, producing terrain-aware place nodes and hierarchically organized regions for outdoor environments. Our method generates a task-agnostic metric-semantic sparse map and constructs a 3DSG from this map for downstream planning tasks, all while remaining lightweight for autonomous robotic operation. Our thorough evaluation demonstrates our 3DSG method performs on par with state-of-the-art camera-based 3DSG methods in object retrieval and surpasses them in region classification while remaining memory efficient. We demonstrate its effectiveness in diverse robotic tasks of object retrieval and region monitoring in both simulation and real-world environments. 
Our Github: \url{https://github.com/BYU-FROST-Lab/Terra}.

\end{abstract}

\begin{figure}[t]
  \centering
  \includegraphics[width=0.99\columnwidth]{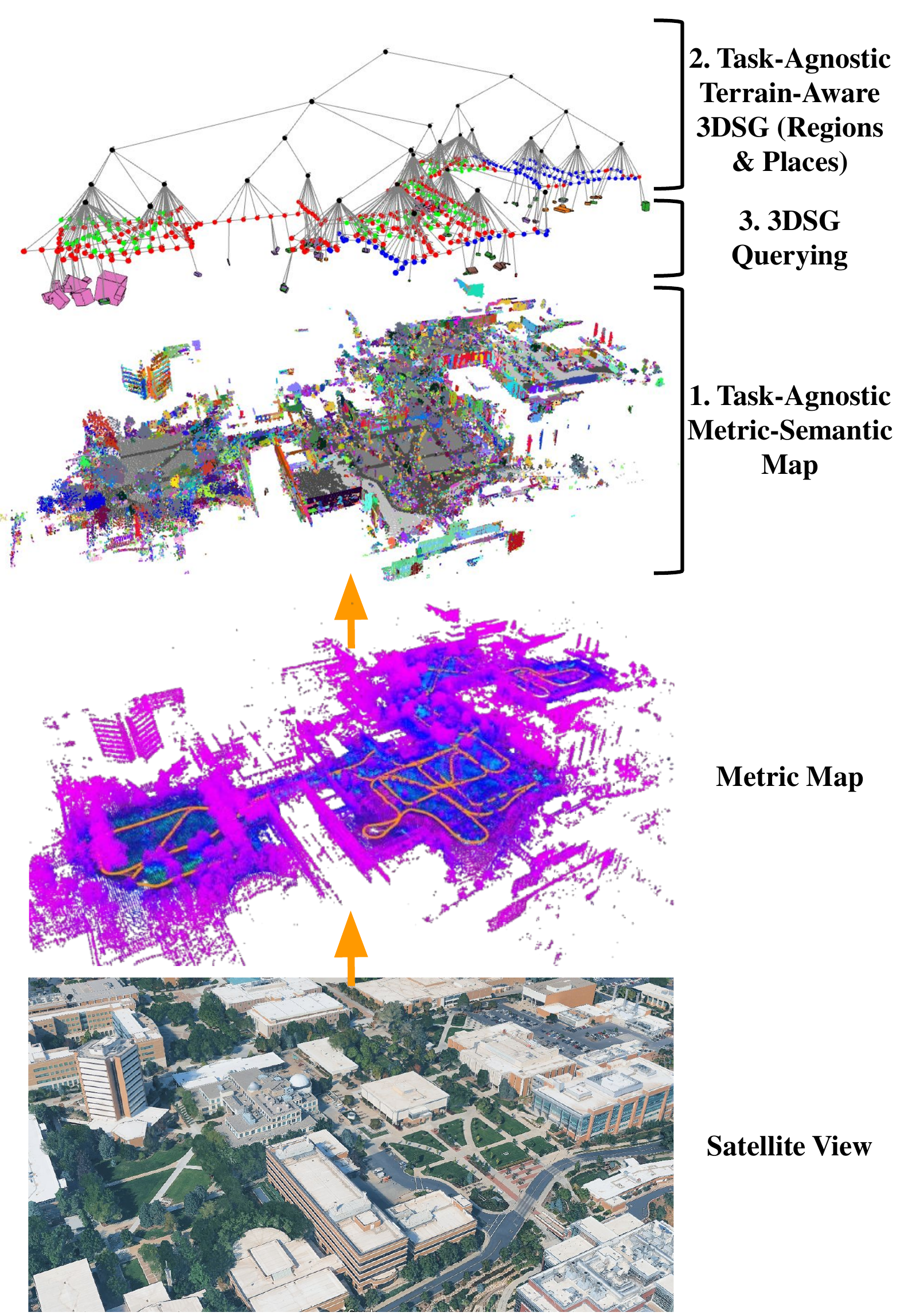}
  \caption{Visual breakdown of our proposed three-phase Terra 3DSG approach built on our large south campus dataset.}
  \label{fig:3dsg_thumbnail}
  \vspace{-0.6cm}
\end{figure}

\section{Introduction}
\label{sec:intro}

Autonomous robotic systems within large-scale outdoor environments have the potential to address a wide range of fundamental societal problems, including search and rescue, forest fires, food delivery, and others.
However, such robotic systems require the ability to robustly and reliably localize within, map, and interpret outdoor scenes at large scales. In this context, we focus on the development of metric-semantic mapping techniques that enable large-scale autonomy in outdoor scenes.

Due to its range, accuracy, and $360^\circ$ field-of-view, LiDAR has become a standard sensor for large-scale outdoor geometric mapping~\cite{slam_legoloam2018, slam_liosam2020shan, slam_xuFASTLIO2Fast2022}. Beyond geometry, some approaches train models to semantically classify points in LiDAR scans~\cite{semlidar_sirohiEfficientLPSEfficient2022, osemlidar_chakravarthyLidarPanoptic2024, osemlidar_liu2023segment, osemlidar_osep2024SAL}. While promising, these models are still largely closed-set (restricted to a fixed set of semantic classes).

Cameras, in contrast, provide rich visual data and have achieved remarkable success in semantic scene understanding. Recent techniques enable more general open-set scene understanding and even enable grounding of visual data with natural language through vision-language models (VLMs)~\cite{maggio2024Clio,werby23hovsg,guConceptGraphsOpenVocabulary2024}. 

Over the last several years, 3D scene graphs (3DSGs) have emerged as a structured approach to build semantically and hierarchically organized metric-semantic maps. Many 3DSG-based methods utilize VLMs and large language models (LLMs) for both scene graph construction and autonomous task planning. Most existing 3DSG methods focus on indoor environments using camera imagery and depth data to build a semantically-classified mesh that forms the base layer of the 3DSG. However, constructing mesh maps over large areas is both computationally and memory intensive. Additionally, camera-derived depth has limited range ($\le20$ meters). These both restrict indoor 3DSG techniques from scaling to large outdoor settings.

In this work, we combine indoor 3DSG techniques with geometrically robust outdoor LiDAR SLAM methods, enabling metric-semantic mapping in large-scale outdoor environments, we term our method Terra (see Fig.~\ref{fig:3dsg_thumbnail}).
We structure the resulting map into a hierarchical scene graph specifically designed to support autonomous outdoor robotic tasks. In particular, since terrain is a key factor for outdoor navigation, we integrate a terrain layer into our 3DSG.
The key contributions of our paper are:
\begin{enumerate}
    \item A novel, memory-efficient, and task-agnostic approach for open-set metric-semantic mapping in large-scale outdoor environments,
    \item A terrain layer in the outdoor 3DSG that supports terrain-aware tasks where VLMs alone struggle,
    \item Hierarchical region layers to handle multiple levels of task abstraction,
    \item An in-depth evaluation on simulated and real world data comparing Terra with state-of-the-art (SOTA) indoor 3DSG methods.
\end{enumerate}

The rest of our paper is outlined as follows. Section \ref{sec:rel_work} provides a brief overview of research in semantic mapping and 3DSG methods. Section \ref{sec:terra_method} gives an explanation for the Terra method. Section \ref{sec:experiments} provides experiments across simulation and real-world datasets demonstrating the capabilities of our Terra method compared to SOTA methods and techniques. Section \ref{sec:conclusion} concludes our findings and explores future work in the area of outdoor 3DSG generation.

\section{Related Work}
\label{sec:rel_work}

\subsection{LiDAR Mapping and Semantics}

Various LiDAR-based SLAM methods are able to robustly build accurate metric maps across diverse, large-scale outdoor scenes~\cite{slam_legoloam2018, slam_liosam2020shan, slam_xuFASTLIO2Fast2022}. Shifting from purely geometric mapping and localization requires the ability to semantically classify individual LiDAR scans.~\citet{semlidar_sirohiEfficientLPSEfficient2022} addressed this with closed-set panoptic segmentation using a neural network. To extend to open-set segmentation,~\citet{osemlidar_chakravarthyLidarPanoptic2024} augmented traditional panoptic semantic segmentation approaches with a class-agnostic clustering method to also group points from unknown categories. More recently,~\citet{osemlidar_liu2023segment} and O\v{s}ep et al.~\cite{osemlidar_osep2024SAL} leveraged vision models to enable open-set LiDAR segmentation.~\citet{osemlidar_liu2023segment} focused on 360-degree LiDAR segmentation given a single class-agnostic segmented image. O\v{s}ep et al.~\cite{osemlidar_osep2024SAL} aligns most closely with our goal of metric-semantic LiDAR mapping using a VLM. However, their approach is limited to per-scan semantic segmentation, does not provide a mapping framework, and the code and models are not publicly available. In this work, we will fuse LiDAR and camera information to build an open-set metric-semantic LiDAR map.

\subsection{Indoor 3D Scene Graphs}

Scene graphs were initially developed to characterize the relationship between different objects in a given image~\cite{krishna2DSG2017}.~\citet{armeni3DScene2019b} extended this concept to 3D scene graphs (3DSGs) that graphically represent the spatial relationship between objects in a 3D scene. This initial work takes a 3D mesh and image semantic classifications and builds a hierarchical representation of the world including a root node, rooms, detected objects, and camera views~\cite{armeni3DSemantic2016}. 

Subsequent approaches refined and expanded this framework. \citet{rosinolKimera2020} introduced Kimera, a full metric-semantic visual SLAM pipeline that takes raw RGB-D images and IMU as input and utilizes modern visual SLAM methods and object classifiers to construct a dense metric-semantic mesh of the environment. Using Kimera for metric-semantic mapping, \citet{rosinol3DDynamic2020} build a 3D dynamic scene graph, similar in structure to~\cite{armeni3DScene2019b}, capable of handling dynamic objects. \citet{hughes2022hydra} later introduced Hydra, which builds upon \cite{rosinolKimera2020,rosinol3DDynamic2020}, but enables real-time SLAM and 3DSG generation. \citet{bavle2025sgraphs20hierarchicalsemantic} introduced situational graphs (S-Graphs), a LiDAR-only 3DSG method that integrates directly with modern factor-graph-based SLAM. While effective for indoor LiDAR scene graph generation, it lacked object classification, which Hydra and its predecessors were capable of. Additionally, each of these methods are limited to closed-set semantic classification.

Naturally, there is interest in extending indoor 3DSG methods from closed-set to open-set semantic scene understanding. Foundation models, such as VLMs, vision foundation models (VFMs) and large language models (LLMs), have made open-set 3DSG methods possible. VLMs in combination with VFMs have been used for open-set object classification~\cite{kassab2024barenecessities,maggio2024Clio,werby23hovsg,guConceptGraphsOpenVocabulary2024}. Commonly in these approaches the VFM provides class-agnostic image segments and the VLM provides a semantic embedding for each segment. Recent work has also utilized LLMs to add in inter- and intra-object relationships to their 3DSG~\cite{guConceptGraphsOpenVocabulary2024}. After 3DSG construction, both VLMs~\cite{maggio2024Clio,werby23hovsg} and LLMs~\cite{guConceptGraphsOpenVocabulary2024} are used for object retrieval and autonomous task execution.   

Most relevant to our work, is Clio~\cite{maggio2024Clio}. \citet{maggio2024Clio} proposed Clio to generate a task-driven 3DSG where the detected objects and region classifications depend on a provided task list. Clio uses FastSAM~\cite{zhao2023fastsam} (VFM) and CLIP~\cite{clip_radfordLearningTransferable2021} (VLM) for image and image segment embeddings. Provided these embeddings, Clio uses an agglomerative information bottleneck (AIB) technique to cluster the segments and place nodes according to task relevance. In this work, \citet{maggio2024Clio} conclude that the optimal 3DSG structure is dependent on the task to be carried out. This insight is of particular relevance as we work to develop 3DSG techniques for modeling unstructured outdoor scenes.

\begin{figure*}[t]
  \centering
  \includegraphics[width=\textwidth]{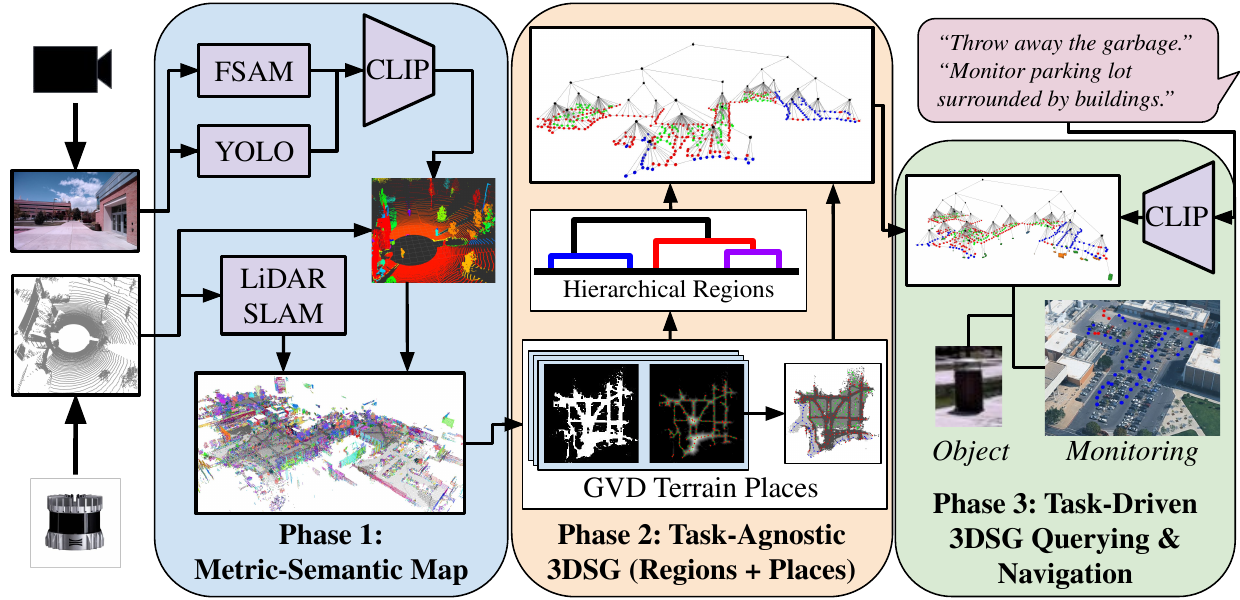}
  \caption{Overview of the three phases that make up our Terra method.}
  \label{fig:terra_method}
  \vspace{-0.5cm}
\end{figure*}

\subsection{Outdoor 3D Scene Graphs}

Significantly less research has focused on building 3DSGs for outdoor environments. This is likely due to the unstructured nature and variability of outdoor environments. 

A few approaches have modified general indoor scene graph methods for outdoor settings. \citet{shanGraph2Nav3D2025} demonstrated capabilities to build an outdoor scene graph using indoor techniques, but lacked true hierarchical semantics and relied on an environment mesh that is computationally intensive to build. \citet{rayTaskMotion2024} adapted Hydra~\cite{hughes2022hydra} for outdoor task and motion planning by redefining place nodes as 2D semantically consistent ground segments. While similar to our proposed terrain-aware technique, their method still relies on mesh reconstruction and closed-set object inclusion, both of which are computationally costly in large outdoor environments. \citet{straderIndoorOutdoor2024} sought to build a more flexible interpretation of region nodes in scene graphs indoors and outdoors. \citet{straderIndoorOutdoor2024} used an LLM-driven ontology to define general object and region relationships and classification for indoor or outdoor environments. This approach generalizes well when key items uniquely identify concepts. For instance, detecting a toilet strongly indicates a bathroom region. However, it struggles with more abstract distinctions, such as differentiating a park from a backyard. 

Recent outdoor-focused 3DSG methods leverage LiDAR and camera data with foundation models for open-set detection, relationship modeling, and task querying. \citet{steinkeCollaborativeDynamic2025b} introduced CURB-OSG, a multi-agent approach for urban environments that handles static and dynamic objects as well as urban features key for navigation in a single 3DSG framework. \citet{zhangParkingSGOpenVocabulary2025} proposed Parking-SG, a parking-lot-specific 3DSG designed for automatic valet parking. While effective in their targeted domains, these methods lack generality across diverse outdoor settings. Moving towards more general outdoor task navigation, \citet{ongATLASNavigator2025} proposed a hierarchical metric-semantic mapping approach, using semantic Gaussian splatting in place of meshes, for task planning. The impressive results in this paper demonstrate the feasibility of a general purpose outdoor metric-semantic map and hierarchical structure to aid in autonomous task completion. The proposed method in this paper most closely aligns with \cite{ongATLASNavigator2025} and \cite{maggio2024Clio}.  

Building on these insights, we introduce an outdoor 3DSG method that combines memory-efficient, task-agnostic, open-set metric-semantic mapping with terrain-aware reasoning and hierarchical region organization. In particular, our method is designed to remain practical for large-scale outdoor robotic operation.

\section{Terra: Hierarchical Terrain-Aware 3D Scene Graphs for Task-Agnostic Outdoor Mapping}
\label{sec:terra_method}

Building on the insights of~\cite{maggio2024Clio}, we believe that 3DSGs are best structured around the requirements of a specific task. However, task-driven 3DSGs are limited in their ability to generalize as task requirements and goals may change. Ideally, a single map of a scene should be applicable to any task, even if tasks are initially unknown.

As such, we propose to break the scene graph generation process into the following phases: (1) task-agnostic metric-semantic mapping, (2) task-agnostic terrain-aware 3DSG construction, and (3) task-driven 3DSG querying and navigation (see Fig.~\ref{fig:terra_method}). The separation of task-agnostic and task-driven mapping enables reuse of the same metric-semantic map across multiple tasks, while allowing task-driven 3DSG querying for each task. Our method currently completes each phase in sequence; however, its design also makes iterative execution possible, opening the door to real-time performance in future work.

\subsection{Phase 1: Task-Agnostic Metric-Semantic Mapping}

We first construct a general task-agnostic metric-semantic map for the scene that will form the backbone for future 3DSG generation. To accomplish this, we must develop a model of the environment that includes task- and class-agnostic semantic information embedded into a geometric representation of the scene. 

\subsubsection{Task-Agnostic Semantic Feature Extraction}
\label{sec:taskagn_semantic_features}
Task-agnostic semantic feature extraction is performed on each time-synchronized RGB image. We first embed terrain features followed by class-agnostic features. In our experiments we found that traditional VLMs struggled to robustly identify terrain (see Table~\ref{tab:terrain_comparison}). This is likely due to the fact that online image captions (on which VLMs like CLIP~\cite{clip_radfordLearningTransferable2021} are trained) rarely describe the terrain type and instead focus on the foreground object of interest.  Therefore, we fine-tune a YOLOv11-segment network~\cite{yolov8_ultralytics} to both mask and classify each terrain type common in our test environments: sidewalk, grass, and asphalt. The terrain class names are embedded using CLIP to produce a latent feature vector for each terrain-type. These vectors are then associated with each terrain mask. FastSAM~\cite{zhao2023fastsam} is also used to generate class-agnostic masks on each image. After removing masks that overlap with the YOLO terrain masks, CLIP produces a latent feature vector for each remaining FastSAM mask.

\subsubsection{LiDAR Point and Semantic Feature Association}
To associate LiDAR points with the image latent feature vectors, we back-project the time-synchronized LiDAR scan into the RGB image frame using precomputed intrinsic and extrinsic calibrations. LiDAR points are then associated with embeddings corresponding to the image segments they appear in. 

\subsubsection{Metric Localization and Mapping} 
LiDAR and IMU data are used to build our sparse geometric global map. We utilize LIO-SAM~\cite{slam_liosam2020shan}, a factor-graph based SLAM system with loop-closures whose global map is a sparse voxelized point cloud.

\subsubsection{Merging Semantic Features into the Sparse Global Point Cloud}
We merge local semantics into the global sparse map by transforming each LiDAR scan into the global frame and using KD-Trees~\cite{bentley1975kdtrees} to associate dense scan points with sparse global points, storing semantic embeddings at each global point. To reduce noise from 2D FastSAM masks, where LiDAR points may include background through gaps or boundaries, we apply DBSCAN~\cite{dbscan_ester1996density} and assign the mask’s CLIP embedding only to the largest point cluster. As new embeddings are incrementally acquired from LiDAR-image pairs, we maintain a tensor of embeddings and track how often each has been assigned to each global point. We retain a new embedding only if its cosine similarity with all existing embeddings is below $0.9$ to limit redundancy and memory use.

\subsection{Phase 2: Task-Agnostic 3DSG Construction}

Provided with the metric-semantic map from the previous phase, we build the task-agnostic place and region node layers of the Terra 3DSG. 

\subsubsection{Terrain-Aware Place Nodes}

The terrain is encoded as a generalized Voronoi diagram (GVD) using the brushfire algorithm~\cite{lauEfficientgridbased2013}, following Hydra's~\cite{hughes2022hydra} adaptation for place node construction. Unlike Hydra, however, our GVD is 2D and there is a separate GVD for each terrain-type. Finally, we connect edge nodes--those with only one neighbor--to the nearest node of a different terrain type, producing a connected places graph. 

Each place node has two semantic embeddings. The first is the terrain class name embedding from CLIP, as described in~\ref{sec:taskagn_semantic_features}. The second embedding builds on the work of~\cite{maggio2024Clio} extending it to outdoor settings. \citet{maggio2024Clio} created place embeddings based on the average CLIP embeddings for all images whose field-of-view included the location of that place node. This works well indoors, where walls and doors obstruct the view of place nodes in different rooms or hallways. However, in an open outdoor setting, this is less effective.  We instead average the CLIP embeddings of all images that were taken within $20$~m of the place node where the place node was in view. In the rare case that there is no image within $20$~m of the place node, we just choose the closest image embedding that had the place node in view.

\subsubsection{Hierarchical Region Nodes}

\begin{figure}[t!]
    \centering
    \begin{subfigure}[b]{0.49\columnwidth}
         \centering
         \includegraphics[width=\columnwidth]{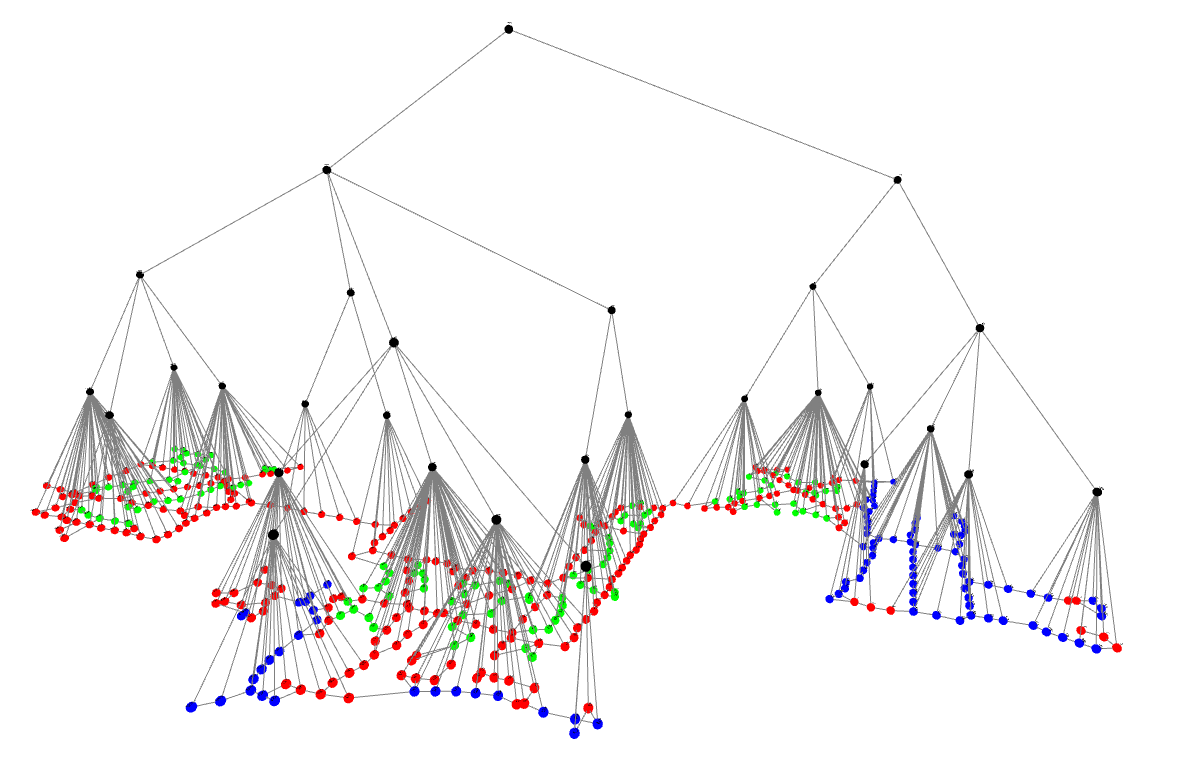}
         \captionsetup{skip=5pt}
         \caption{}
         \label{fig:aggcluster}
    \end{subfigure}
    \begin{subfigure}[b]{0.49\columnwidth}
         \centering
         \includegraphics[width=\columnwidth]{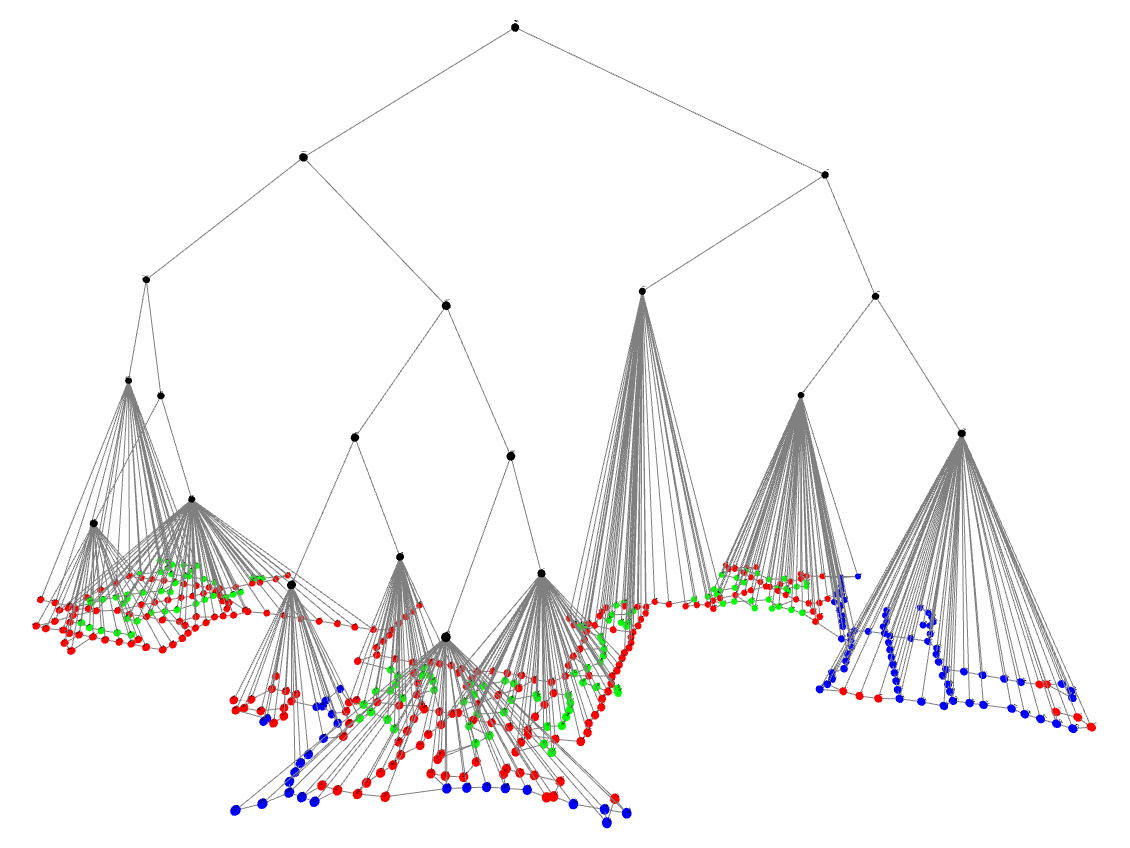}
         \caption{}
         \label{fig:speccluster}
    \end{subfigure}
    \caption{Visual comparison of region clustering techniques on campus dataset: (a) agglomerative and (b) spectral.}
    \vspace{-0.5cm}
    \label{fig:region_clustering}
\end{figure}

The benefits of a truly hierarchical 3DSG come when the hierarchy enables broader levels of environment abstraction to aid in abstract task completion. We propose that clustering and averaging the place node image embeddings incrementally to form multiple hierarchical layers of region nodes, provides multiple layers of semantic environment abstraction. This concept is most closely related to~\cite{ongATLASNavigator2025}, who apply agglomerative clustering over semantic Gaussians to build a three-level object-hierarchy. Whereas their work emphasizes object-centric semantic hierarchies, we aim to capture region-level abstractions, leading us to cluster place nodes rather than our metric-semantic map. We develop two different approaches for creating hierarchical region nodes: 1) agglomerative clustering and 2) spectral clustering. 

For agglomerative clustering, we add edges to the places graph to make it a fully connected graph. The edge weights follow~\cite{ongATLASNavigator2025} to combine both semantic and geometric distances between nodes. We convert the fully connected graph into a distance matrix and perform agglomerative clustering. We split the hierarchical clusters at four different levels based on distance thresholds of $50,100,200,400$~m producing a four-level hierarchical graph. The embeddings for each new region node is the average of all CLIP embeddings of its respective children nodes.

Our spectral clustering approach focuses reversely on splitting the graph of place nodes into two using the same edge weights as in agglomerative clustering to build an affinity matrix. This is repeated recursively for each cluster until the cluster is below a semantic difference or size threshold. The semantic difference of a cluster is the difference of cosine-similarity scores of the most and least similar node to the cluster. The size of a cluster is based on the 2D bounding box area of that cluster. The resulting region hierarchies from both approaches are shown for a campus dataset in Fig.~\ref{fig:region_clustering}.

\subsection{Phase 3: Task-Driven 3DSG Querying and Navigation}

Based on the output of Phases 1 and 2, our Terra method can combine information from the metric-semantic map, places, and regions layers to accomplish a wide-range of outdoor autonomous tasks. In this paper, we focus on two types of outdoor tasks including object retrieval and region monitoring. Terrain-aware path planning for object retrieval tasks is then explained.  

\subsubsection{Object Retrieval}
\label{sec:obj_retrieval}

In the object retrieval task, the goal is to identify relevant 3D object bounding boxes in the environment, given a natural language query. We explore two strategies. The first relies only on the metric-semantic map. In this approach, we locate all global points most similar to the object query above a threshold $\alpha$, cluster neighboring points with DBSCAN to form a bounding box, and then add the box to the 3DSG by linking it to its closest place node (referred to as {\em Terra MS} in section \ref{sec:experiments_openset_obj_eval}). The second strategy leverages the full Terra 3DSG prior (including the region and place hierarchy) before conducting DBSCAN clustering. In this approach, we search for region nodes whose semantic similarity to the query exceeds the similarity threshold, $\alpha$. We repeat this process with the selected region's child place nodes, and then perform the first strategy, Terra-MS, on this subset of the global point cloud (referred to as {\em Terra 3DSG} in section \ref{sec:experiments_openset_obj_eval}).    

\subsubsection{Region Monitoring}
\label{sec:region_monitoring}

For monitoring tasks, the goal is to identify the task-relevant region that needs to be monitored by the robot. Given a monitoring task natural language query, we select the top-$k$ relevant region nodes. We then select all child place nodes of the chosen region nodes above the task similarity threshold $\alpha$. These chosen place nodes represent the region for the robot to monitor. 

\subsubsection{Navigation}
\label{sec:navigation}

Navigation is performed on the 3DSG places layer, where place nodes and connections derived from the GVD ensure maximally safe path trajectories. Given a start node (the robot's map location) and a task, we apply Terra 3DSG object detection to locate task-relevant objects and select the associated goal node. An A* planner then finds the shortest path between the start and goal nodes. Terrain preferences are encoded as large edge weights, steering A* away from undesirable terrain nodes. Prohibited terrain types (e.g., water for a ground robot) are assigned infinite edge weights, forcing the planner to avoid them entirely.

\subsection{Platform details}
All real-world experiments used a modified wheelchair platform. The wheelchair is equipped with an OAK-D LR camera for RGB imagery data and an Ouster OS1-128 LiDAR with an integrated IMU for performing SLAM. Both are mounted on top of the wheelchair frame.

\section{Experiments}
\label{sec:experiments}

\subsection{Datasets}

\subsubsection{Simulation}

\begin{figure}[t]
  \centering
  \vspace{-0.5cm}
  \includegraphics[width=0.95\columnwidth]{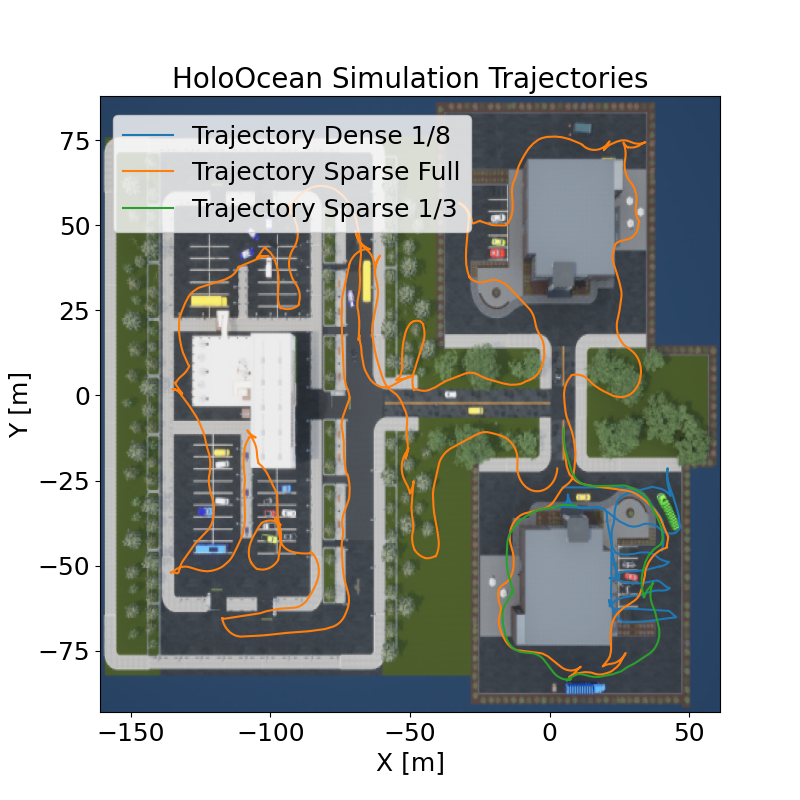}
  \vspace{-0.3cm}
  \caption{Birds-eye-view of the simulated HoloOcean environment and the three different trajectories taken to collect the three simulation datasets.}
  \vspace{-0.5cm}
  \label{fig:holoocean_sim}
\end{figure}

Our photorealistic simulated business campus environment was designed in the HoloOcean simulator~\cite{Potokar22icra} (Fig.~\ref{fig:holoocean_sim}). Experiments in the environment used a forward-facing RGB camera, depth camera, IMU, and LiDAR with parameters matching real-world sensors. We collected three datasets: \textbf{Dense $1/8$ Map}--multiple passes around a small parking lot covering one-eighth of the campus (375.9~m trajectory, blue in Fig.~\ref{fig:holoocean_sim}); \textbf{Sparse $1/3$ Map}--a single loop around one building covering one-third of the campus (228.0~m, green); \textbf{Sparse Full Map}--a single traversal of the entire environment (1563.6~m, orange). We fine-tuned a YOLOv11 model to detect terrain classes: ``sidewalk" [red], ``grass" [green], and ``asphalt" [blue]. The dataset contained 1420 training and 406 validation image-label pairs.

\subsubsection{Real-world}
For real-world terrain and qualitative experiments, we collected a dataset of the southern portion of our university campus with RGB, LiDAR and IMU sensor data. We fine-tuned a YOLOv11 model to detect terrain classes: ``sidewalk" [red], ``grass" [green], and ``asphalt" [blue]. The dataset contained 11,176 training and 3,193 validation image-label pairs.

For both simulated and real-world experiments, the robot was manually driven in each scene to enable data collection and Terra 3DSG construction.
Following which, object querying, region monitoring, and path planning experiments were executed and evaluated on the constructed Terra 3DSG.
This allows us to isolate and evaluate the capabilities of the proposed mapping and scene graph representation independently of the navigation
stack.

\subsection{2D Terrain Classification}

To motivate our custom YOLO-v11-seg model for terrain classification, we evaluated several SOTA VLM-based approaches on a dataset of 374 terrain-labeled campus images with varied lighting and shadow conditions. The evaluated terrain-types were sidewalk, grass, and asphalt. While environment-specific models are expected to outperform general-purpose VLMs, our goal was to assess how well VLMs handle open-set terrain segmentation. The methods were: FastSAM + CLIP, generating masks with FastSAM and assigning CLIP embeddings to masks, keeping masks with cosine similarity above $0.26$; Grounding-DINO + FastSAM, where terrain prompts generate bounding boxes via Grounding-DINO, and FastSAM matches the mask with the greatest IoU; CLIP-DINOiser, assigning CLIP embeddings to image patches with the greatest cosine similarity; YOLOE-v11s-seg, outputting segment masks given terrain prompts; and our custom YOLO-v11n-seg, trained to segment sidewalk, grass, and asphalt. Optimal VLM prompts where determined empirically: \textit{sidewalk}, \textit{dirt or dead grass or grass}, \textit{asphalt}. Performance was measured using mean IoU (mIoU), frequency weighted mIoU (F-mIoU), F1-score, and FPS (see Table~\ref{tab:terrain_comparison}) using an Intel Core Ultra 9 285K CPU, NVIDIA GeForce RTX 5070 GPU and 32 GB of RAM.

\begin{table}[t]
\centering
\begin{tabular}{lllll}
\hline
Method & mIoU $\uparrow$ & F-mIoU $\uparrow$ & F1 $\uparrow$ & FPS $\uparrow$ \\
\hline
FastSAM + CLIP      & 0.257  & 0.186 & 0.378 & 1.18 \\
CLIP-DINOiser~\cite{clipdinoiser2024}       & 0.498  & 0.157 & 0.593 & 4.72 \\
G-DINO~\cite{gdino2024} + FastSAM    & \underline{0.568}  & 0.159 & \underline{0.624} & 5.41 \\
YOLOE-v11s-seg~\cite{wang2025yoloerealtimeseeing}     & 0.512  & \underline{0.314} & 0.611 & \underline{102} \\
YOLO-v11n-seg (Ours)       & \textbf{0.786}  & \textbf{0.451} & \textbf{0.854} & \textbf{119} \\
\hline
\end{tabular}
\caption{Comparison of object detection and segmentation methods for terrain segmentation across 374 labeled campus images. Demonstrates the robust capabilities of our YOLO terrain model filling this terrain gap of SOTA VFM and VLM models.}
\label{tab:terrain_comparison}
\vspace{-0.3cm}
\end{table}

Table~\ref{tab:terrain_comparison} highlights the need for terrain-specific segmentation models, as all general-purpose methods achieve relatively low frequency-weighted mIoU scores. This metric is important because most of the collected imagery data contains sidewalks, which are easier to segment than more complex natural terrain (e.g. grass). Runtime is also critical for real-time applications. YOLO-v11n-seg achieves best overall performance across all metrics, but if a custom terrain model is unavailable, YOLOE~\cite{wang2025yoloerealtimeseeing} offers the best efficiency, while Grounding-DINO~\cite{gdino2024} provides the highest accuracy among general-purpose approaches. The parameters of each model were initially qualitatively adjusted by visual inspection of the outputs. After identifying the parameters with the greatest influence on performance, the models were evaluated on the data set to further refine the parameters using quantitative metrics (such as IoU, precision, and recall). 





\subsection{Open-set 3D Object Retrieval}
\label{sec:experiments_openset_obj_eval}

\begin{table}[t]
\centering
\resizebox{\columnwidth}{!}{%
\begin{tabular}{lllllllll}
\hline
Test & Method & IoU $\uparrow$ & SAcc $\uparrow$ & RAcc $\uparrow$ & SPrec $\uparrow$ & RPrec $\uparrow$ & F1 $\uparrow$ \\
\hline
\multirow{6}{*}{\rotatebox{90}{\shortstack{Dense \\ 1/8 Map}}} 
& Clio-Prim*        & 0.010  & 0.000 & 0.000 & 0.003 & 0.003 & 0.000 \\
& Clio-Online*      & 0.024  & \textbf{0.125} & \underline{0.125} & 0.111 & 0.111 & 0.118 \\
& Clio-Batch*       & -  & - & - & - & - & - \\
& Terra MS-Avg   & \textbf{0.064}  & \textbf{0.125} & \underline{0.125} & \textbf{0.400} & \textbf{0.400} & \underline{0.190} \\
& Terra MS-Max   & \underline{0.058}  & \textbf{0.125} & \textbf{0.312} & 0.091 & \underline{0.273} & \textbf{0.291} \\
& Terra 3DSG & \textbf{0.064} & \textbf{0.125} & \underline{0.125} & \underline{0.250} & 0.250 & 0.167 \\
\hline
\multirow{6}{*}{\rotatebox{90}{\shortstack{Sparse \\ 1/3 Map}}} 
& Clio-Prim*        & 0.036  & \textbf{0.214} & \textbf{0.214} & 0.036 & 0.036 & 0.061 \\
& Clio-Online*        & \underline{0.067}  & \textbf{0.214} & \textbf{0.214} & \textbf{0.333} & \underline{0.333} & \underline{0.261} \\
& Clio-Batch*        & 0.018  & 0.071 & \underline{0.143} & 0.143 & 0.143 & 0.143 \\
& Terra MS-Avg   & 0.060  & 0.071 & \underline{0.143} & 0.111 & 0.222 & 0.174 \\
& Terra MS-Max   & \textbf{0.070} & \underline{0.143} & \textbf{0.214} & \underline{0.286} & \textbf{0.429} & \textbf{0.286} \\
& Terra 3DSG & \textbf{0.070}  & 0.071 & \underline{0.143} & 0.077 & 0.154 & 0.148 \\
\hline
\multirow{6}{*}{\rotatebox{90}{\shortstack{Sparse \\ Full Map}}}
& Clio-Prim*        & -  & - & - & - & - & - \\
& Clio-Online*    & -  & - & - & - & - & - \\
& Clio-Batch*    & -  & - & - & - & - & - \\
& Terra MS-Avg* & \underline{0.067} & \underline{0.176} & \underline{0.176} & 0.189 & 0.189 & 0.182 \\
& Terra MS-Max* & \textbf{0.080} & 0.162 & 0.162 & \textbf{0.236} & \textbf{0.236} & \underline{0.192} \\
& Terra 3DSG* & 0.066 & \textbf{0.189} & \textbf{0.189} & \underline{0.200} & \underline{0.200} & \textbf{0.194} \\
\hline
\end{tabular}
} 
\caption{Evaluation of open-set object detection on three simulated datasets using simple object queries, ``Find the \{object\}." Dense $1/8$: 16 objects in 5 queries. Sparse $1/3$: 14 objects in 7 queries. Sparse Full: 74 objects in 10 queries. ``-" indicates method failure due to RAM or clustering failures. Metrics show Terra performs on par with SOTA mesh-based 3DSG (Clio) and scales to large scenes.}
\label{tab:exp2_obj_ret}
\vspace{-0.5cm}
\end{table}

We evaluate the open-set object detection performance of our sparse outdoor point cloud method against Clio~\cite{maggio2024Clio}, a SOTA indoor image-based 3DSG approach. Following definitions in~\cite{maggio2024Clio}, we report IoU, strict/relaxed accuracy, strict/relaxed precision, and F1 scores based on ground-truth and predicted oriented bounding boxes. The two Terra MS variants differ in how CLIP embeddings are assigned to metric-semantic points: Terra MS-Avg uses the average of all associated embeddings, while Terra MS-Max assigns the most frequently occurring embedding. Terra tests were on an Intel Core i7-10850H CPU, NVIDIA Quadro RTX 3000 Mobile GPU and 32 GB of RAM laptop. Clio and the object-retrieval portion of Terra in the sparse full map test, denoted by *, were run on an Intel Xeon Gold 6242 CPU, NVIDIA  Quadro 6000 GPU and 128 GB of RAM.

The results for these experiments are recorded in Table~\ref{tab:exp2_obj_ret}. In general the results appear lower than similar metrics in indoor scenes. This is likely due to the fact that this is a simulation and in outdoor operations, most objects are only partially observed from a distance via a single pass, while in most indoor 3DSG datasets care is taken to observe all objects multiple times from various different views. For the dense test, Terra outperforms Clio. This can partially be attributed to the failure of the mesh reconstruction SLAM methods that Clio is built upon. Our method performs on par with Clio in the sparse $1/3$ test demonstrating the semantic effectiveness of our sparse point cloud approach compared to dense mesh reconstruction methods. Lastly, Terra is able to run on the large sparse full map dataset when Clio fails due to dataset size, demonstrating Terra's large-scale mapping capabilities. We find that our Terra MS approaches for object detection remains optimal across varying dataset sizes and object queries.

\begin{figure}[t]
    \centering
    \includegraphics[width=\columnwidth]{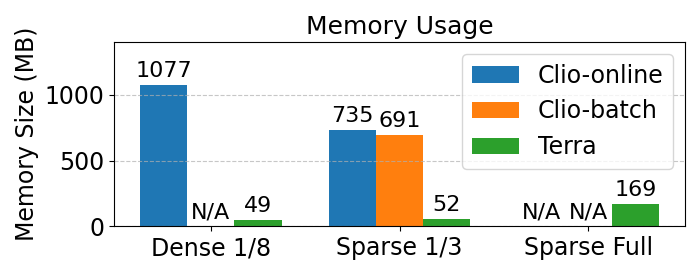}
    \vspace{-0.65cm}
    \caption{Memory usage of three methods across simulated experiments. N/A indicates failure due to RAM or clustering failures. Terra 3DSG remains more memory-efficient than all Clio variants, even on the largest dataset.}
    \label{fig:memory_usage}
    \vspace{-0.6cm}
\end{figure}

Figure~\ref{fig:memory_usage} presents the memory usage of the saved 3DSG files between the two methods. For Clio we report the 
file size of Clio's saved 3DSG, \texttt{dsg.json}. For Terra the reported size is the graph, embeddings, and full metric-semantic point cloud. The differences can be attributed to the individual segment embeddings and dense places layer of Clio since it was designed for smaller indoor settings.

\begin{figure}
  \centering
  \includegraphics[width=\columnwidth]{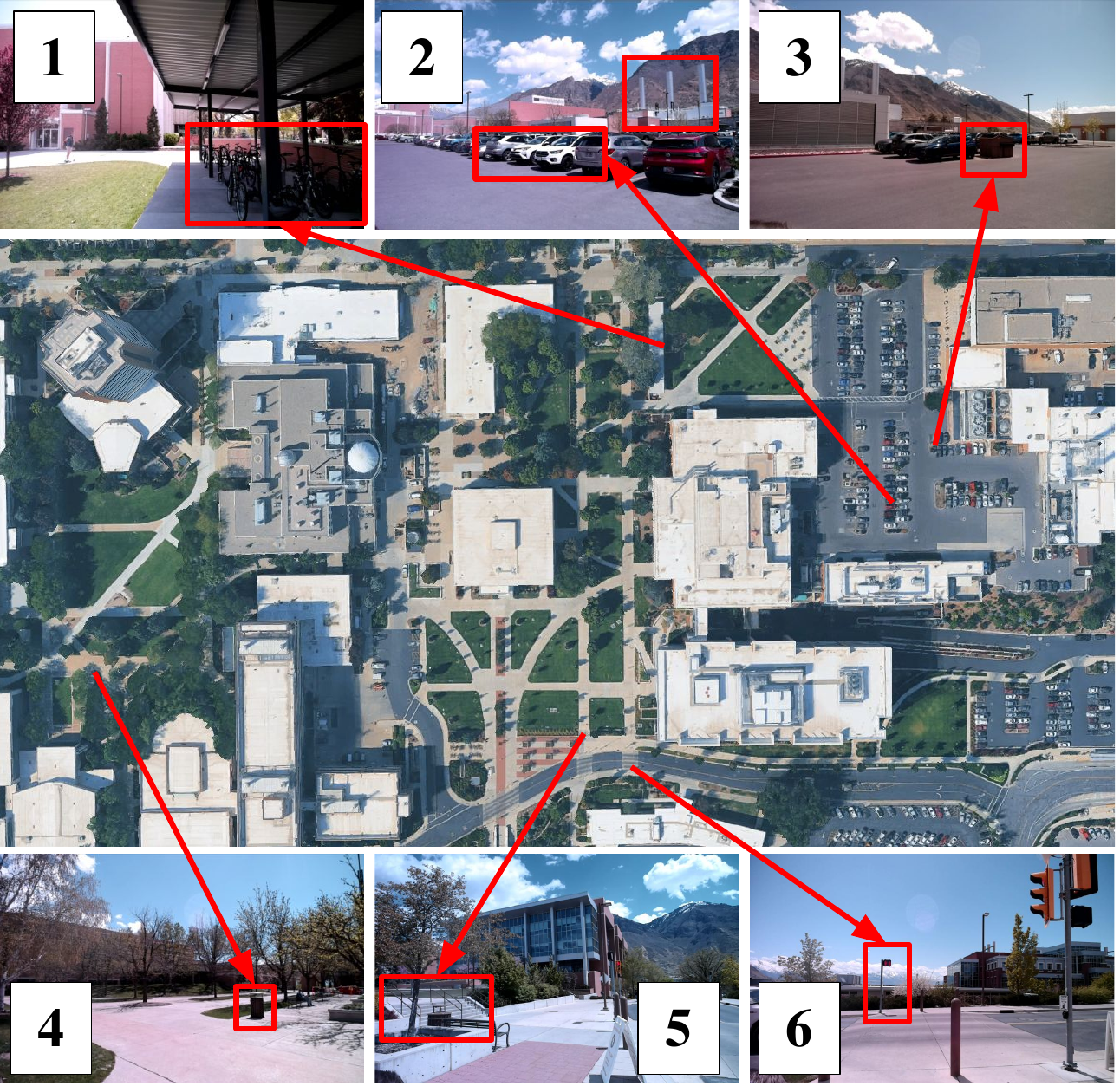}
  \caption{Real-world object detection. Sample images of objects detected across the six tasks. This demonstrates successful object detection across a variety of diverse tasks.}
  \label{fig:realworld_obj_retrieval}
\end{figure}

We also conducted real-world tests for object retrieval. For these tests we use Terra agglomerative-regions with the 3DSG searching technique and a similarity threshold of $\alpha = 0.25$. Figure~\ref{fig:realworld_obj_retrieval} shows images of objects returned when providing the following campus map related queries: 1) “Get bike from the bicycle rack,” 2) “Find my white car by the smoke stacks,” 3) “Empty trash in dumpster,” 4) “Throw away garbage in garbage can,” 5) “Avoid the stairs,” and 6) “Cross the crosswalk by the stoplight.” These results demonstrate that Terra can be used to enable outdoor complex object retrieval.

\subsection{Hierarchical Region Evaluation}
\label{sec:exp_hier_reg_eval}

\begin{table}[t]
\centering
\resizebox{\columnwidth}{!}{%
\begin{tabular}{lllll}
\hline
Method & Precision $\uparrow$ & Recall $\uparrow$ & F1 $\uparrow$ & Runtime (ms) $\downarrow$ \\
\hline
AIB~\cite{maggio2024Clio} on Terra Places-layer   & 0.179  & 0.134 & 0.150 & 755 \\
Terra Spectral      & \textbf{0.531} & 0.462 & 0.468 & \textbf{62} \\
Terra Agglomerative & 0.411 & \textbf{0.583} & \textbf{0.471} & 91 \\
\hline
\end{tabular}
} 
\caption{Region evaluation. Evaluating place node clustering accuracy based on region task. Demonstrates that for outdoor settings, hierarchical place node clustering aligns more with human intuition than the AIB-clustering approach.}
\label{tab:exp3_regions}
\vspace{-0.5cm}
\end{table}

\begin{figure}[h]
    \centering
    \begin{subfigure}{0.48\columnwidth}
        \includegraphics[width=\linewidth]{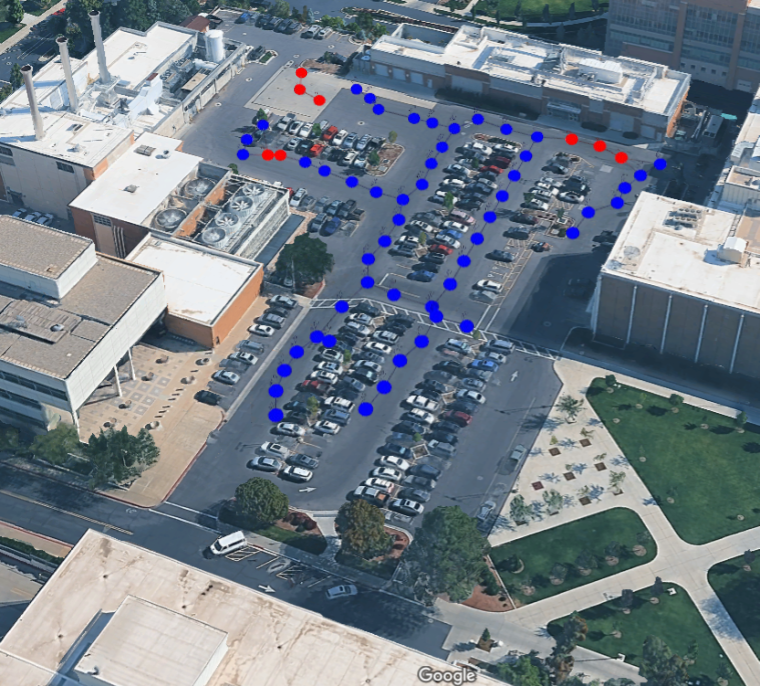}
        \caption{Query: ``Large parking lot surrounded by buildings"}
    \end{subfigure}
    \hfill
    \begin{subfigure}{0.48\columnwidth}
        \includegraphics[width=0.725\linewidth]{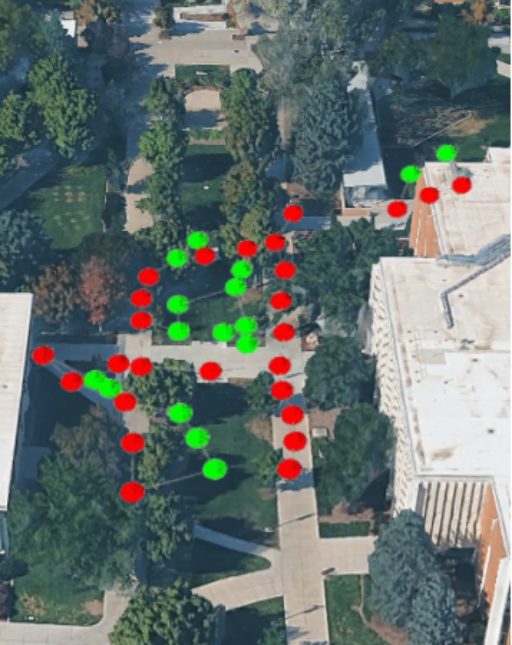}
        \caption{Query: ``Two long parallel sidewalks lined by trees"}
    \end{subfigure}
    \caption{Terra spectral predicted output for two of the four provided queries. Predicted place nodes overlapped on respective Google map screenshot. Visually displaying place node alignment with human intuition of our region method.}
\label{fig:qual_regions}
\vspace{-0.5cm}
\end{figure}

The benefit of hierarchical regions are evaluated on real-world data both quantitatively and qualitatively on the Intel i7 laptop. We had a graduate, familiar with campus, label four unique regions of campus based on the satellite image in Fig.~\ref{fig:realworld_obj_retrieval} and describe each with a single sentence. These descriptions were used as the region monitoring task prompts to predict which place nodes correlate to each prompt. The Terra spectral and agglomerative techniques are described in~\ref{sec:region_monitoring}. We implemented the batch AIB-method in Python presented in~\cite{maggio2024Clio} which Clio uses for task-driven place node clustering into regions. The precision, recall, and F1 metrics report consistent region clustering for both our spectral and agglomerative approaches. AIB struggles in outdoor region clustering likely due to the ambiguity of region definitions in outdoor environments. We demonstrate the qualitative results from two of the four region queries in Fig.~\ref{fig:qual_regions}.

\subsection{Navigation}

\begin{figure*}[t!]
    \centering
    \begin{subfigure}[b]{0.32\textwidth}
         \centering
         \includegraphics[width=\textwidth]{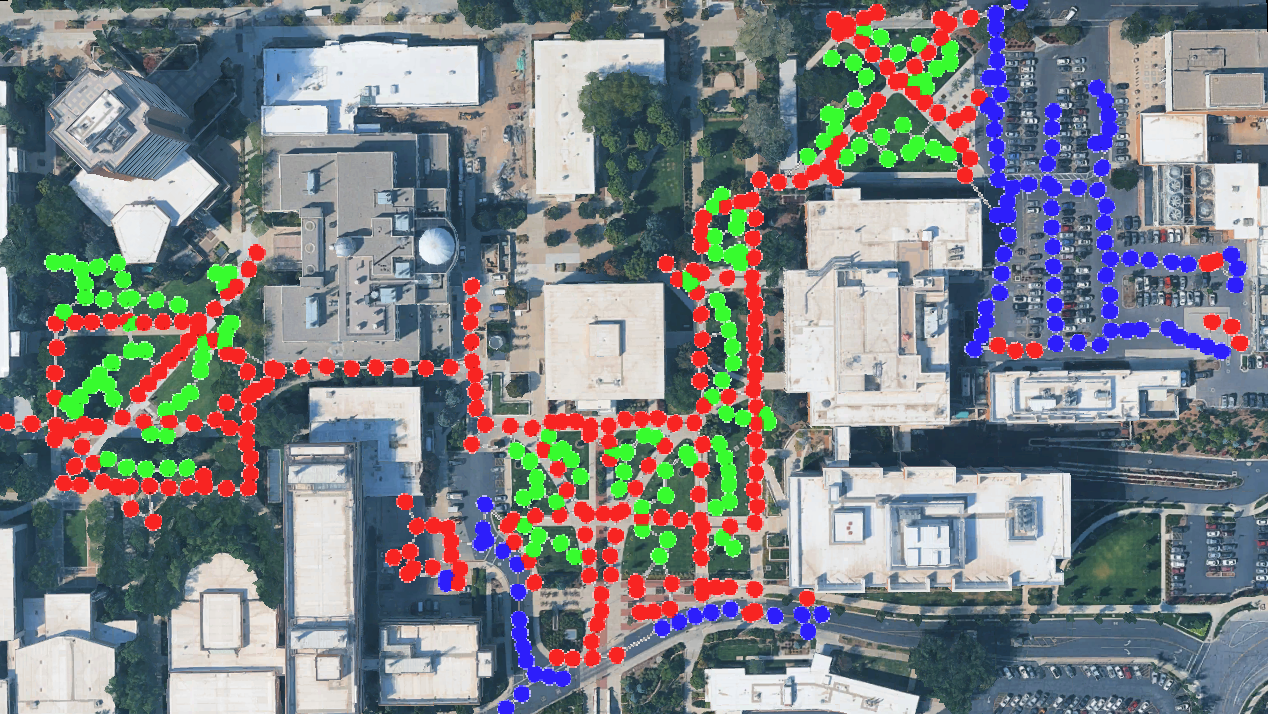}
         \captionsetup{skip=5pt}
         \caption{}
         \label{fig:terrainplaces}
    \end{subfigure}
    \begin{subfigure}[b]{0.325\textwidth}
         \centering
         \includegraphics[width=\textwidth]{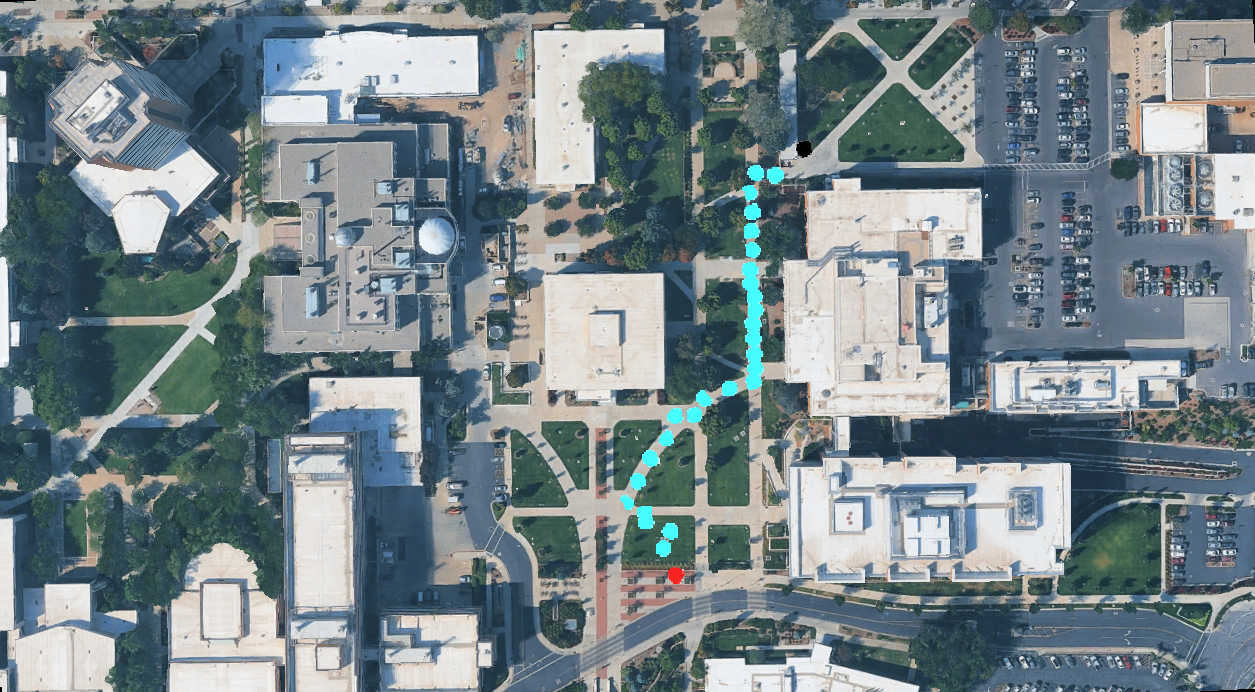}
         \caption{}
         \label{fig:pathplan_noterr}
    \end{subfigure}
    \begin{subfigure}[b]{0.335\textwidth}
         \centering
         \includegraphics[width=\textwidth]{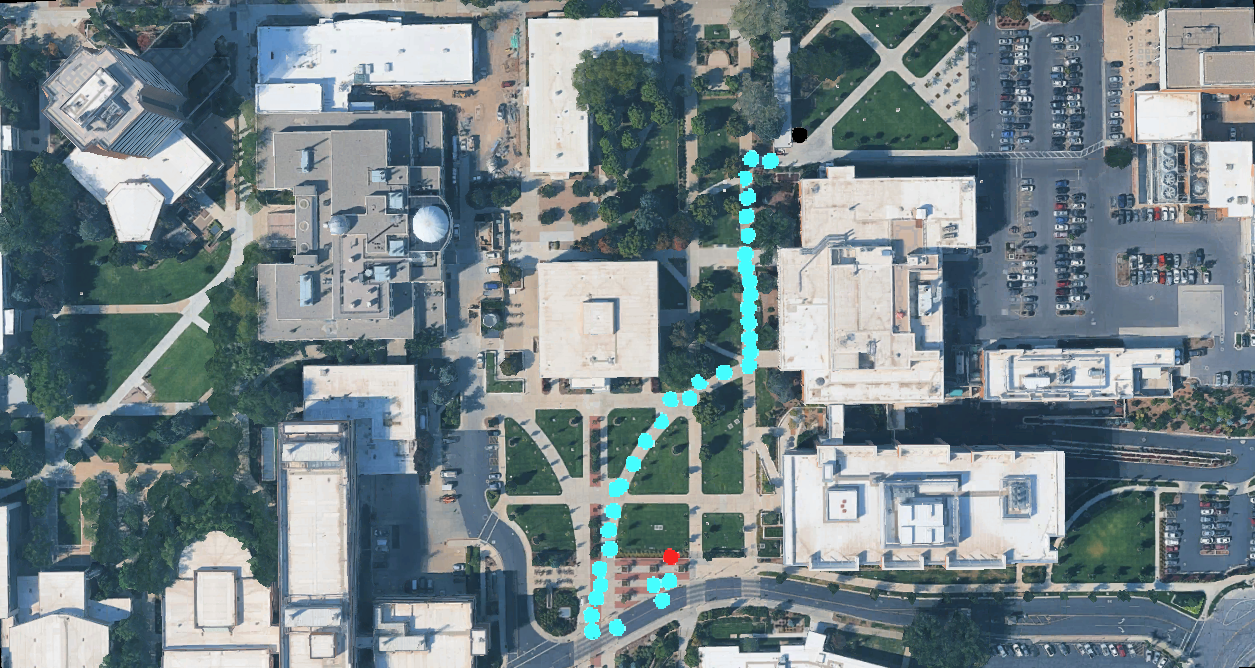}
         \caption{}
         \label{fig:pathplan_sidewalk}
    \end{subfigure}
    \caption{Path planning results for the task ``Get bike from the bicycle rack." (a) Terrain-colored places layer. (b) Path from start (red) to goal (black). (c) Same start and goal but with sidewalk preference, modifying initial green grass path nodes, illustrating terrain-aware planning for task completion.}
    \vspace{-0.75cm}
    \label{fig:pathplanning}
\end{figure*}

We lastly demonstrate terrain-aware navigation using our Terra-3DSG method with agglomerative clustering and similarity threshold of $0.25$. Following Sec.~\ref{sec:navigation}, we start at the red node, shown in Fig.~\ref{fig:pathplan_noterr} and Fig.~\ref{fig:pathplan_sidewalk}, and provide the task ``Get bike from the bicycle rack." The task-relevant goal node is shown in black in Fig.~\ref{fig:pathplan_noterr} and Fig.~\ref{fig:pathplan_sidewalk}. In Fig.~\ref{fig:pathplan_noterr}, A* computes the shortest path (cyan nodes) without terrain preferences. In Fig.~\ref{fig:pathplan_sidewalk}, we add a preference for sidewalk terrain rerouting the path to avoid the initial three green grass terrain nodes, and instead go through one asphalt node and stay on the sidewalk for the rest. These results highlight Terra's capability for terrain-aware path planning. All run on the Intel i7 laptop.

\section{Conclusion}
\label{sec:conclusion}

We have proposed a general, sparse, task-agnostic metric-semantic mapping and 3DSG approach for large-scale outdoor environments. We have shown the need for terrain-capable models to aid in terrain understanding in diverse outdoor settings. We have demonstrated both quantitatively and qualitatively the effectiveness of this approach in various object retrieval and monitoring tasks. For future work, we are interested in extending this method into a full real-time pipeline and exploring more robust 3DSG techniques for off-road and complex natural environments. While we leverage CLIP for its efficiency, compact size, and task-agnostic open-set embeddings, its reliance on prompt-tuning and view-dependent recognition remain limitations, which future LLM-based VLMs may help overcome.

\textbf{Acknowledgments:} ChatGPT-4 and 5 AI models were used as collaborative tools for code development. All generated code was thoroughly edited and tested by the authors.


\bibliographystyle{style/IEEEtranN}
{\footnotesize

}

\end{document}